\documentclass[review,11pt]{ReportTemplate}
\usepackage{bm}
\usepackage{graphicx}
\usepackage{algorithm}
\usepackage{algorithmic}
\usepackage{rotating}
\usepackage{amssymb}
\usepackage{booktabs}
\usepackage{multirow}
\usepackage{booktabs}

\newcommand{\bs}{\boldsymbol}
\newcommand{\x}{\mathbf{x}}
\newcommand{\w}{\mathbf{w}}
\newcommand{\X}{\mathcal {X}}
\newcommand{\Y}{\mathcal {Y}}

\newcommand{\caperf}{{CAPO}}
\newcommand{\caperfone}{{CAPO}$_1$}
\newcommand{\caperffive}{{CAPO}$_5$}
\newcommand{\svmperf}{{SVM$^\text{perf}$}}
\newcommand{\svmperflin}{{SVM$^\text{perf}_\text{lin}$}}
\newcommand{\svmperfrbf}{{SVM$^\text{perf}_\text{rbf}$}}
\newcommand{\svmlight}{{SVM$^\text{light}$}}
\newcommand{\svmlightlin}{{SVM$^\text{light}_\text{lin}$}}
\newcommand{\svmlightrbf}{{SVM$^\text{light}_\text{rbf}$}}

\newtheorem{prop}{Proposition}

\newenvironment{proof}{\textbf{Proof:}\ }{\hspace{\stretch{1}}$\square$}

\begin{document}
\begin{frontmatter}
\title{Efficient Optimization of Performance Measures \\ by Classifier Adaptation}

\author{Nan Li$^{1,2}$}
\author{Ivor W. Tsang$^{3}$}
\author{Zhi-Hua Zhou$^1$\corref{cor1}}
\address{$^1$National Key Laboratory for Novel Software Technology, Nanjing University, Nanjing 210046, China \\ 
$^3$School of Mathematical Scicences, Soochow University, Suzhou 215006, China \\
$^2$School of Computer Engineering, Nanyang Technological University, 639798, Singapore } 
\cortext[cor1]{\small Corresponding author. Email: zhouzh@lamda.nju.edu.cn}

\begin{abstract}
In practical applications, machine learning algorithms are often needed to learn classifiers that optimize domain specific performance measures. Previously, the research has focused on learning the needed classifier in isolation, yet learning nonlinear classifier for nonlinear and nonsmooth performance measures is still hard. In this paper, rather than learning the needed classifier by optimizing specific performance measure directly, we circumvent this problem by proposing a novel two-step approach called as CAPO, namely to first train nonlinear auxiliary classifiers with existing learning methods, and then to adapt auxiliary classifiers for specific performance measures. In the first step, auxiliary classifiers can be obtained efficiently by taking off-the-shelf learning algorithms. For the second step, we show that the classifier adaptation problem can be reduced to a quadratic program problem, which is similar to linear \svmperf and can be efficiently solved. By exploiting nonlinear auxiliary classifiers, CAPO can generate nonlinear classifier which optimizes a large variety of performance measures including all the performance measure based on the contingency table and AUC, whilst keeping high computational efficiency. Empirical studies show that CAPO is effective and of high computational efficiency, and even it is more efficient than linear \svmperf.
\end{abstract}


\end{frontmatter}

\section{Introduction}
In real-world applications, different user requirements often employ different domain specific performance measures to evaluate the success of learning algorithms. For example, {F1-score} and {Precision-Recall Breakeven Point} (PRBEP) are usually employed in text classification; {Precision} and {Recall} are often used in information retrieval; {Area Under the ROC Curve} (AUC) and {Mean Average Precision} (MAP) are important to ranking. Ideally, to achieve good prediction performance, learning algorithms should train classifiers by optimizing the concerned performance measures. However, this is usually not easy due to the nonlinear and nonsmooth nature of many performance measures like {F1-score} and {PRBEP}.

During the past decade, many algorithms have been developed to optimize frequently used performance measures, and they have shown better performance than conventional methods~\cite{Lafferty-SIGIR-01,CaruanaNCK04,Lafferty-AISTAT-05,Joachims05,CaoIR06,BurgesRL06}. By now, the research has focused on training the needed classifier in isolation. But, in general, it is still challenging to design general-purpose learning algorithms to train nonlinear classifiers optimizing nonlinear and nonsmooth performance measures, though it is very needed in practice. For example, \svmperf~proposed by Joachims~\cite{Joachims05} can efficiently optimize a large variety of performance measures in the linear case, but its nonlinear kernelized extension suffers from computational problems~\cite{YuJ08,JoachimsY09}.

In this paper, rather than directly designing sophisticated algorithms to optimize specific performance measures, we take a different strategy and present a novel two-step approach called CAPO to cope with this problem. Specifically, we first train auxiliary classifiers by exploiting existing off-the-shelf learning algorithms, and then adapt the obtained auxiliary classifiers to optimize the concerned performance measure.
Note that in the literature, there have been proposed many algorithms that can train the auxiliary classifiers quite efficiently, even on large-scale data, thus the first step can be easily performed.
For the second step, to make use of the auxiliary classifiers, we consider the classifier adaptation problem under the function-level adaptation framework~\cite{Yang08}, and formulate it as a quadratic program problem which is similar to linear \svmperf~\cite{Joachims05} and can also be efficiently solved. Hence, in total, CAPO can work efficiently.

A prominent advantage of CAPO is that it is a flexible framework, which can handle different types of auxiliary classifiers and a large variety of performance measures including all the performance measure based on the contingency table and AUC.
By exploiting nonlinear auxiliary classifiers, CAPO can train nonlinear classifiers optimizing the concerned performance measure with low computational cost. This is very helpful, because nonlinear classifiers are preferred in many real-world applications but training such a nonlinear classifier is often of high computational cost (e.g. nonlinear kernelized \svmperf).
In empirical studies, we perform experiments on  data sets from different domains. It is found that CAPO is more effective and more efficient than state-of-the-art methods, also it scales well with respect to training data size and is robust with the parameters. It is worth mentioning that the classifier adaptation procedure of CAPO is even more efficient than linear \svmperf, though it employs the same cutting-plane algorithm to solve the classifier adaptation problem.

The rest of this paper is organized as follows. Section~\ref{sec:background} briefly describes some background, including the problem studied here and \svmperf. Section~\ref{sec:caperf} presents our proposed \caperf~approach. Section~\ref{sec:dis} gives some discussions on related work. Section~\ref{sec:exp} reports on our empirical studies, followed by the conclusion in Section~\ref{sec:conclusion}.

\section{Optimizing Performance Measures}\label{sec:background}
In this section, we first present the problem of optimizing performance measures, and then introduce \svmperf~\cite{Joachims05} and its kernelized extension.

\subsection{Preliminaries and Background}
In machine learning tasks, given a set of $n$ training examples ${D} = \{(\x_1, y_1),$ $\ldots, (\x_n, y_n) \}$, where $\x_i \in \X$ and $y_i \in \{-1, +1\}$ are input pattern and its class label, our goal is to learn a classifier $f(\x)$ that minimizes the expected risk on new data sample $S = \{(\x_1', y_1'),$ $\ldots, (\x_m', y_m') \}$, i.e., 
$$
R^\Delta(f) = \mathbb{E}_{S}[\Delta((y_1', \ldots, y_m'), (f(\x_1'), \ldots, f(\x_m')))] \ ,
$$
where $\Delta((y_1', \ldots, y_m'), (f(\x_1'), \ldots, f(\x_m')))$ is the loss function which quantifies the loss of $f$ on $S$. Subsequently, we use the notation $\Delta(f; S)$ to denote $\Delta((y_1', \ldots, y_m'), (f(\x_1'), \ldots, f(\x_m')))$ for convenience.  Since it is intractable to compute the expectation $\mathbb{E}_{S}[\cdot]$, discriminative learning methods usually approximate the expected risk $R^\Delta(f)$ using the empirical risk
$$
\hat{R}_D^\Delta(f)  = \Delta(f;D) \ ,
$$
which measures $f(\x)$'s loss on the training data $D$, and then train classifiers by minimizing empirical risk or regularized risk.
In practice, domain specific performance measures are usually employed to evaluate the success of learnt classifiers. Thus, good performance can be expected if the classifiers are trained by directly optimizing  the concerned performance measures. Here, we are interested in regarding the loss function $\Delta$ as practical performance measures (e.g., F1-score and PRBEP), instead of some kinds of surrogate functions (e.g., hinge loss and exponential loss). In this situation, the loss function $\Delta$ can be nonlinear and nonsmooth function of training examples in $D$, thus it is computationally challenging to optimize the empirical risk $\Delta$ in practice.

In the literature, some methods have been developed to optimize frequently-used performance measures, such as AUC~\cite{Ferri02,HerschtalICML04}, F1-score~\cite{MusicantKO03}, NDCG and MAP~\cite{YueIR07,XuIR08,ValizadeganNIPS09}.
Among existing methods that try to optimize performance measures directly, the \svmperf~proposed by Joachims~\cite{Joachims05} is a representative example. One of its attractive advantages is that by employing the multivariate prediction framework, it can directly handle a large variety of performance measures, including AUC and all measures that can be computed from the contingency table, while most of other methods are specially designed for one specific performance measure. Subsequently, we describe it and also show its limitation.

\subsection{\svmperf and Its Kernelized Extension}\label{sec:svmperf}
Since many performance measures cannot be decomposed over individual predictions,  \svmperf~\cite{Joachims05} takes a multivariate prediction formulation and considers to map a tuple of $n$ patterns $\bar{\x} = (\x_1, \ldots, \x_n)$ to a tuple of $n$ class labels $\bar{y} = (y_1, \ldots, y_n)$ by
$$
\bar{f}: \X^n \mapsto \Y^n \ ,
$$
where $\Y^n \subseteq \{-1, +1\}^n$ is set of all admissible label vectors. To implement this mapping, it exploits a discriminant function and makes prediction as
\begin{eqnarray}\label{eq:argmax}
\bar{f}(\bar{\x}) = \mathop{\arg\max}_{\bar{y}'\in \Y^n} ~ \w^\top \Psi(\bar{\x}, \bar{y}) \ ,
\end{eqnarray}
where $\w$ is a parameter vector and $\Psi(\bar{\x}, \bar{y}')$ is a feature vector relating $\bar{\x}$ and $\bar{y}'$. Obviously, the computational efficiency of the inference (\ref{eq:argmax}) highly depends on the form of the feature vector $\Psi(\bar{\x}, \bar{y}')$ .

\subsubsection{Linear Case}

\begin{algorithm}[!t]
\caption{Cutting-plane algorithm for training linear \svmperf~\cite{Joachims05}}\label{algo:cpa}
\begin{algorithmic}[1]
\STATE {Input:} ${D} = \{(\x_i, y_i)\}_{i=1}^{n}$, $C$, $\epsilon$\\
\STATE $\mathcal{W} \leftarrow \emptyset $\\
\REPEAT
\STATE $(\w, \xi) \leftarrow \mathop{\arg\min}_{\w, \xi \geq 0} ~~ \frac{1}{2}\|\w\|^2 + C\xi$\\
\quad\quad\quad\quad$\text{s.t.} ~~  \forall~\bar{y}' \in \mathcal{W}: $ ~ $\w^\top [\Psi(\bar{\x}, \bar{y}) - \Psi(\bar{\x}, \bar{y}')] \geq \Delta(\bar{y}, \bar{y}') - \xi $ \ ,
\STATE find the most violated constraint by ~$\bar{y}' \leftarrow \mathop{\arg\max}_{\bar{y}'' \in \Y^n} \{ \Delta(\bar{y}, \bar{y}'') + \w^\top \Psi(\bar{\x}, \bar{y}'') \}$\\
\STATE $\mathcal{W} \leftarrow \mathcal{W} \cup \{ \bar{y}'\}$
\UNTIL{$\Delta(\bar{y}, \bar{y}') - \w^\top [\Psi(\bar{\x}, \bar{y}) - \Psi(\bar{\x}, \bar{y}')] \leq \xi + \epsilon$}
\end{algorithmic}
\end{algorithm}

In~\cite{Joachims05}, the feature vector $\Psi(\bar{\x}, \bar{y}')$ is restricted to be
$$
\Psi(\bar{\x}, \bar{y}') = \sum_{i=1}^n y_i' \x_i \ ,
$$
thus the argmax in (\ref{eq:argmax}) can be achieved by assigning $y_i'$ to $\mathop{\rm sign}(\w^\top\x_i)$, leading to a linear classifier
$
f(\x) = \mathop{\rm sign}[\w^\top\x]
$.
To learn the parameter $\w$, the following optimization problem is formulated
\begin{eqnarray}\label{eqn-svmperf}
\min_{\w, \xi \geq 0} && \frac{1}{2}\|\w\|^2 + C~\xi\\
\text{s.t.} &&  \forall~\bar{y}' \in \Y^n \setminus \bar{y}: ~~
 \w^\top [\Psi(\bar{\x}, \bar{y}) - \Psi(\bar{\x}, \bar{y}')] \geq \Delta(\bar{y}, \bar{y}') - \xi \nonumber \ ,
\end{eqnarray}
where $\Delta(\bar{y}, \bar{y}')$ is the loss of mapping $\bar{\x}$ to $\bar{y}'$ while its true label vector is $\bar{y}$. It is not hard to find that $\Delta(\bar{y}, \bar{y}')$ can incorporate many types of performance measures, and the problem (\ref{eqn-svmperf}) optimizes an upper bound of the empirical risk~\cite{Joachims05}.

While there are a huge number of constraints in (\ref{eqn-svmperf}), the cutting-plane algorithm in Algorithm~\ref{algo:cpa} can be used to solve it, and this algorithm has been shown to need at most $O(1/\epsilon)$ iterations to converge to an $\epsilon$-accurate solution~\cite{Joachims05,Joachims09a}.
In each iteration, it needs to find the most violated constraint by solving
\begin{eqnarray}\label{eq:find-most}
\mathop{\arg\max}_{\bar{y}' \in \Y^n} ~ \{ \Delta(\bar{y}, \bar{y}') + \w^\top \Psi(\bar{\x}, \bar{y}') \} \ .
\end{eqnarray}
It has been shown that if the discriminant function $\w^\top \Psi(\bar{\x}, \bar{y}')$ can be written in the form $\sum_{i=1}^n y_i' f(\x_i)$, the inference (\ref{eq:find-most}) can be solved for many performance measures in polynomial time, that is, $O(n^2)$ for contingency table based performance measures (such as F1-score) and $O(n\log n)$ for AUC~\cite{Joachims05}.
Hence, Algorithm~\ref{algo:cpa} can train \svmperf~in polynomial time.

\subsubsection{Kernelized Extension}
Using kernel trick, the linear \svmperf~described above can be extended to the non-linear case~\cite{Joachims09a}. It is easy to obtain that the dual of (\ref{eqn-svmperf}) as
\begin{eqnarray}\label{eqn-svmperf-dual}
\max_{\bs{\alpha} \geq 0} && -\frac{1}{2} \bs{\alpha}^\top \mathbf{H} \bs{\alpha} + \sum_{\bar{y}'\in\Y^n} \alpha_{\bar{y}'} \Delta(\bar{y}, \bar{y}') \\
\text{s.t.} &&  \sum_{\bar{y}'\in\Y^n} \alpha_{\bar{y}'}= C \ ,  \nonumber
\end{eqnarray}
where $\bs{\alpha}$ is the column vector of $\alpha_{\bar{y}'}$'s and $\mathbf{H}$ is the Gram matrix with the entry $\mathbf{H}(\bar{y}', \bar{y}'')$ as
$$
\mathbf{H}(\bar{y}', \bar{y}'') = \left[\Psi(\bar{\x},  \bar{y}) - \Psi(\bar{\x}, \bar{y}')]^\top[\Psi(\bar{\x}, \bar{y}) - \Psi(\bar{\x}, \bar{y}'')\right] \ .
$$
By replacing the primal problem with its dual in Line 4, it is easy to get the dual variant of Algorithm~\ref{algo:cpa}, which can solve the problem (\ref{eqn-svmperf-dual}) in at most $O(1/\epsilon)$ iterations~\cite{Joachims09a,JoachimsY09}. In the solution, each $\alpha_{\bar{y}'}$ corresponds to a constraint in $\mathcal{W}$,
and the discriminant function $\w^\top \Psi(\bar{\x}, \bar{y}')$ in (\ref{eq:argmax}) can be written as
$$
\w^\top \Psi(\bar{\x}, \bar{y}') = \sum_{\bar{y}'' \in \mathcal{W}} \alpha_{\bar{y}''} [\Psi(\bar{\x}, \bar{y}) - \Psi(\bar{\x}, \bar{y}'')]^\top\Psi(\bar{\x}, \bar{y}') \ .
$$
Obviously, the inner product $\Psi(\bar{\x}, \bar{y}')^\top\Psi(\bar{\x}, \bar{y}'')$ can be computed via a kernel $K(\bar{\x}, \bar{y}', \bar{\x}, \bar{y}'')$. However, if so, it can be found that the argmax in (\ref{eq:argmax}) and (\ref{eq:find-most}) will become computationally intractable. Hence, feature vectors of the following form are used
$$
\Psi(\bar{\x}, \bar{y}') = \sum_{i=1}^n y_i' \Phi(\x_i) \ ,
$$
where $\Phi(\x_i)^\top\Phi(\x_j)$ can be computed via a kernel function $K(\x_i, \x_j) = \Phi(\x_i)^\top\Phi(\x_j)$. Then, the discriminant function becomes
\begin{eqnarray}\label{eq:df-kernel}
\w^\top \Psi(\bar{\x}, \bar{y}') = \sum_{i=1}^n y_i'  \sum_{j=1}^n \beta_j K(\x_i, \x_j) \ ,
\end{eqnarray}
where $\beta_j = \sum_{\bar{y}'' \in \mathcal{W}} \alpha_{\bar{y}''}(y_j - y_j'')$. In this case, the argmax in (\ref{eq:argmax}) can be achieved by assigning each $y_i'$ with $\mathop{\rm sign}\left[\sum_{j=1}^n \beta_j K(\x_i, \x_j)\right]$, which produces the kernelized classifier
$$
f(\x) = \mathop{\rm sign}\left[\sum_{i=1}^n \beta_i K(\x, \x_i)\right]  .
$$

However, in each iteration, the Gram matrix $\mathbf{H}$ needs to be updated by adding a new row/column for the new constraint. Suppose $\bar{y}^+$ is added, for every $\bar{y}' \in \mathcal{W}$, it requires computing
$$
\mathbf{H}(\bar{y}', \bar{y}^+) = \sum_{i=1}^n\sum_{j=1}^n (y_i - y_i')(y_j - y_j^+) K(\x_i, \x_j) \ .
$$
Thus, let $m$ denote the number of constraints in $\mathcal{W}$ and $n$ denote the data size, it takes $O(m n^2)$ kernel evaluations in each iteration. Also, it should be noted that computing the discriminative function (\ref{eq:df-kernel}) also requires $O(n^2)$ kernel evaluations, and this adds to the computational cost of the inference (\ref{eq:find-most}).
These issues make the kernelized extension of \svmperf~suffer from computational problems, even on reasonably-sized data set.
However, as we know, nonlinear classifiers are quite needed in many practical application. Hence, training nonlinear classifier that optimizes a specific performance measure becomes central to this work.

\section{Classifier Adaptation for Performance Measures}\label{sec:caperf}
In this section, we introduce our proposed approach CAPO, which is short for {Classifier Adaptation for Performance measures Optimization}. 

\subsection{Motivation and Basic Idea}
Notice the fact that it is generally not straightforward to design learning algorithms which optimize specific performance measure, while there has been many well-developed learning algorithms in the literature and some of them can train complex nonlinear classifiers quite efficiently. Our intuitive motivation of this work is to exploit these existing algorithms to help training the needed classifier that optimizes the concerned performance measure.

Specifically, denote $f^*(\x)$ as the ideal classifier which minimizes the empirical risk $\Delta(f; D)$, it is generally not easy to design algorithms which can efficiently find $f^*(\x)$ in the function space by minimizing $\Delta(f; D)$ due to its nonlinear and nonsmooth nature, especially when we are interested in complex nonlinear classifiers.
Meanwhile, by using many off-the-shelf learning algorithms, we can get certain classifier $f'(\x)$ quite efficiently, even on large-scale data set. Obviously, $f'(\x)$ can differ from the ideal classifier $f^*(\x)$, since it may optimize a different loss from $\Delta(f; D)$. However, since many performance measures are closely related, for example, both F1-score and PRBEP are functions of precision and recall, the average AUC is an increasing function of accuracy~\cite{CortesNIPS03}, $f'(\x)$ can be regarded as a rough estimated classifier of $f^*(\x)$, then we conjecture that $f'(\x)$ will be helpful to finding  $f^*(\x)$ in the function space, for example, it can reduce the computational cost of searching the whole function space.
Subsequently, $f'(\x)$ is called as auxiliary classifier and $f^*(\x)$ as target classifier.

To implement this motivation, we take classifier adaptation techniques \cite{LiB07,Yang07} which have achieved successes in domain adaptation~\cite{Daume-06}.
Specifically, after getting the auxiliary classifier $f'(\x)$, we adapt it to a new classifier $f(\x)$ and it is expected that the adapted classifier $f(\x)$ can achieve good performance in terms of the concerned performance measure. For the classifier adaptation procedure, it is expected that
\begin{itemize}
  \item The adapted classifier outperforms auxiliary classifier in terms of concerned performance measure;
  \item The adaptation procedure is more efficient than directly training a new classifier for concerned performance measure;
  \item The adaptation framework can handle different types of auxiliary classifiers and different performance measures.
\end{itemize}
Since many existing algorithms can train auxiliary classifiers efficiently, we focus on the classifier adaptation procedure in the remainder of the paper.

\subsection{Classifier Adaptation Procedure}
For the aim of this work, we study the classifier adaptation problem under the function-level adaptation framework, which is originally proposed for domain adaption in~\cite{Yang07,Yang08}.

\subsubsection{Single Auxiliary Classifier}
The basic idea is to directly modify the decision function of auxiliary classifier which can be of any type. Concretely, given one auxiliary classifier $f'(\x)$, we construct the new classifier $f(\x)$ by adding a delta function $f_\delta(\x) = \w^\top \Phi(\x)$, i.e.,
$$
f(\x) 
      = \mathop{\rm sign}\left[f'(\x) + \w^\top \Phi(\x)\right] \ ,
$$
where $\w$ is the parameter of $f_\delta(\x)$, and $\Phi(\cdot)$ is a feature mapping. It should be noted that $f'(\x)$ is the auxiliary classifier directly producing +1/-1 predictions, and it can be of any type (e.g., SVM, neural network, decision tree, etc) because it is treated as a ``black-box'' in CAPO; while $f_\delta(\x)$ is a real-valued function, which is added to modify the decision of $f'(\x)$ such that $f(\x)$ can achieve good performance in terms of our concerned performance measure. Obviously, our task is reduced to learn the delta function $f_\delta(\x)$, and hence the classifier $f(\x)$.

Based on the principle of regularized risk minimization, it should consider the problem
\begin{eqnarray}\label{eq:rrm}
    \min_{\w}~ \Omega(\w) + C \cdot \Delta(\bar{y}, \bar{y}^*) \ ,
\end{eqnarray}
where $\Omega(\w)$ is a regularization term, $\Delta(\bar{y}, \bar{y}')$ is the empirical risk on training data $D$ with $\bar{y} = (y_1, \ldots, y_n)$ are the true class labels and $\bar{y}^* = (f(\x_1), \ldots, f(\x_n))$ are the predictions of $f(\x)$, and $C$ is the regularization parameter.
In practice, the problem (\ref{eq:rrm}) is not easy to solve, mainly due to the following two issues:
 \begin{enumerate}
   \item For some multivariate performance measures like F1-score, the empirical risk $\Delta$ cannot be decomposed over individual predictions, i.e., they cannot be written in the form of $\Delta(\bar{y}, \bar{y}') = \sum_{i=1}^n \ell(y_i, h(\x_i))$ ;
   \item The empirical risk $\Delta$ can be nonconvex and nonsmooth;
 \end{enumerate}
To cope with these issues, inspired by \svmperf~\cite{Joachims05}, we take the multivariate prediction formulation. That is, instead of learning $f(\x): \X \mapsto \Y$ directly, we consider $\bar{f}: \X^n \mapsto \Y^n$ which maps a tuple of $n$ patterns $\bar{\x} = (\x_1, \ldots, \x_n)$ to $n$ class labels $\bar{y} = (y_1, \ldots, y_n)$. Specifically, the mapping is implemented by maximizing a discriminant function $F(\bar{\x}, \bar{y})$, i.e.,
\begin{eqnarray}\label{eq:camax}
\bar{y} = \mathop{\arg\max}_{\bar{y}' \in \Y^n} F(\bar{\x}, \bar{y}') \ .
\end{eqnarray}
In this work, $F(\bar{\x}, \bar{y}) = \sum_{i=1}^n y_i f(\x_i)$ is used, so the argmax in (\ref{eq:camax}) can be easily obtained by assigning $y_i'$ with $f(\x)$. In this way, (\ref{eq:camax}) becomes
$$
\bar{y} = \mathop{\arg\max}_{\bar{y}' \in \Y^n} \begin{bmatrix} 1 \\ \w \end{bmatrix}^\top \Upsilon(\bar{\x}, \bar{y}), \ ~~\text{where}~~~
\Upsilon(\bar{\x}, \bar{y}) = \sum_{i=1}^n y_i \begin{bmatrix}f'(\x_i)\\ \Phi(\x_i) \end{bmatrix} \ .
$$

Instead of directly minimizing $\Delta(\bar{y}, \bar{y}')$, we consider its convex upper bound as follows.

\begin{prop}
Given training data $D$ and the discriminative funciton $F(\x, \bar{y})$, the risk function
\begin{eqnarray}\label{eqn:cvx-bnd}
R(\w; D) = \max_{\bar{y}' \in \Y^n} \left[F(\bar{\x}, \bar{y}') - F(\bar{\x}, \bar{y}) + \Delta(\bar{y}, \bar{y}')\right]
\end{eqnarray}
is a convex upper bound of the empirical risk $\Delta(\bar{y}, \bar{y}^*)$ with $\bar{y}^* = \mathop{\arg\max}_{\bar{y}' \in \Y^n} F(\bar{\x}, \bar{y})$.
\end{prop}
\vspace{-5mm}
\begin{proof}
The convexity of (\ref{eqn:cvx-bnd}) with respect to $\w$ is due to the fact that $F$ is linear in $\w$ and a maximum of linear functions is convex.
Since $\bar{y}^* = \mathop{\arg\max}_{\bar{y}' \in \Y^n} F(\bar{\x}, \bar{y})$, it follows
$$
R(\w; D) \geq F(\mathbf{x}, \bar{y}^*) - F(\mathbf{x}, \bar{y}) + \Delta(\mathbf{y}, \bar{y}^*)\\
\geq \Delta(\mathbf{y},\bar{y}^*) \ .
$$
Thus, $R(\w; D)$ is a convex upper bound.
\end{proof}

Consequently, by taking $\Omega(\w) = \|\w\|^2$ and the convex upper bound $R(\w; D)$, the problem (\ref{eq:rrm}) becomes
\begin{eqnarray}\label{eqn-OP-single}
\min_{\w, \xi \geq 0} && \frac{1}{2}\|\w\|^2 + C \xi\\
\text{s.t.} &&  \forall~\bar{y}' \in \Y^n \setminus \bar{y}: ~~ \begin{bmatrix} 1 \\ \w \end{bmatrix}^\top [\Upsilon(\bar{\x}, \bar{y}) - \Upsilon(\bar{\x}, \bar{y}')] \geq \Delta(\bar{y}, \bar{y}') - \xi \nonumber \ ,
\end{eqnarray}
where $\xi$ is a slack variable introduced to hide the max in (\ref{eqn:cvx-bnd}).

Although the regularization term $\|\w\|^2$ has the same form as that of \svmperf~in (\ref{eqn-svmperf}), it has a different meaning, as stated in following proposition.

\begin{prop}
By minimizing the regularization term $\|\w\|^2$ in (\ref{eqn-OP-single}), the adapted classifier $f(\x)$ is made to be near the auxiliary classifier $f'(\x)$ in reproducing kernel Hilbert space.
\end{prop}
\vspace{-5mm}
\begin{proof}
The Lagrangian function of (\ref{eqn-OP-single}) is
$$
L = \frac{1}{2}\|\w\|^2 + \left(C -\gamma - \sum_{{\bar{y}'} \in \Y^n}\alpha_{\bar{y}'} \right)\xi -  \sum_{{\bar{y}'} \in \Y^n}\alpha_{\bar{y}'} \left(\begin{bmatrix} 1 \\ \w \end{bmatrix}^\top [\Upsilon(\bar{\x}, \bar{y}) - \Upsilon(\bar{\x}, \bar{y}')] - \Delta(\bar{y}, \bar{y}')\right),
$$
where $\alpha_{\bar{y}'}$ and $\gamma$ are Lagrangian multipliers. By setting the derivative of $L$ with respect to $\w$ to zero, we obtain
$$
\w = \sum_{i=1}^n\beta_i  \Phi(\x_i)\ \quad\quad \text{and} \quad\quad f_\delta(\x) = \sum_{i=1}^n\beta_i K(\x_i, \x) \ ,
$$
where $\beta_i=\sum_{\bar{y}' \in \Y^n} \alpha_{\bar{y}'}(y_i - y_i')$ and $K(\x_i, \x) = \Phi(\x_i)^\top \Phi(\x)$.

Since $f(\x) = f'(\x) + f_\delta(\x)$, the distance between $f$ and $f'$ in RKHS is 
$$
\|f - f'\|^2 = \|f_\delta\|^2 = \langle f_\delta, f_\delta \rangle = \sum_{i=1}^n\sum_{j=1}^n\beta_i \beta_j K(\x_i, \x_j) \ .
$$ 
Meanwhile, since $\w = \sum_{i=1}^n\beta_i  \Phi(\x_i)$, we have
$$
\|\w \|^2 = \sum_{i=1}^n \sum_{j=1}^n \beta_i\beta_i  \Phi(\x_i)^\top\Phi(\x_i) \ .
$$
By computing $\Phi(\x_i)^\top\Phi(\x_i)$ via the kernel $K(\x_i, \x_j)$, we can obtain $\|f - f'\|^2 = \|\w \|^2$, which completes the proof.
\end{proof}

In summary, by solving the problem (\ref{eqn-OP-single}), CAPO finds the adapted classifier $f(\x)$ near the auxiliary classifier $f'(\x)$ such that $f(\x)$ minimizes an upper bound of the empirical risk, and the parameter $C$ balances these two goals.

\subsubsection{Multiple Auxiliary Classifiers}
If there are multiple auxiliary classifiers, rather than choosing one, we learn the target classifier by leveraging all the auxiliary classifiers. A straightforward idea is to construct an ensemble of them, then the ensemble is treated as a single classifier to be adapted. Suppose we have $m$ auxiliary classifiers $f^1(\x), \ldots, f^m(\x)$, the target classifier $f(\x)$ can be formulated as
\begin{eqnarray}\label{eqn-f-target}
f(\x) = \mathop{\rm sign}\left[\sum_{i=1}^{m} a_i f^i(\x) + \w^\top \Phi(\x)\right] \ ,
\end{eqnarray}
where $a_i$ is the weight of the auxiliary classifier $f^i(\x)$, and $f_\delta(\x) = \w^\top \Phi(\x)$ is the delta function as above. We learn the ensemble weights $\mathbf{a} = [a_1, \ldots, a_m]^\top$ and the parameter $\w$ of $f_\delta(\x)$ simultaneously.
Let $\mathbf{f}_i = [f^1(\x_i), \ldots, f^m(\x_i)]^\top$ and
$$
\Psi(\bar{\x}, \bar{y}) = \sum_{i=1}^n y_i \begin{bmatrix} \mathbf{f}_i \\ \Upsilon(\x_i) \end{bmatrix}.
$$
Following the same strategy as above, the following problem is formulated. 
\begin{eqnarray}\label{eqn-OP-multi}
\min_{\mathbf{a}, \w, \xi \geq 0} && \frac{1}{2}\|\w\|^2 + \frac{1}{2} B \|\mathbf{a}\|^2 + C \xi\\
\text{s.t.} &&  \forall~\bar{y}' \in \Y^n \setminus \bar{y}: ~~\begin{bmatrix} \mathbf{a} \\ \w \end{bmatrix}^\top [\Psi(\bar{\x}, \bar{y}) - \Psi(\bar{\x}, \bar{y}')]
        \geq \Delta(\bar{y}, \bar{y}') - \xi \nonumber.
\end{eqnarray}
where $\|\mathbf{a}\|^2$ penalizes large weights on the auxiliary classifiers. It prevents the target classifier $f(\x)$ from too much reliance on the auxiliary classifiers, because they do not directly optimize the target performance measure. The term $\|\w\|^2$ measures the distance between $f(\x)$ and $\sum_{i=1}^{m} a_i f^i(\x)$ in the function space. Thus, minimizing $\frac{1}{2} \|\w\|^2$ finds the final classifier $f(\x)$ near the ensemble of auxiliary classifiers $\sum_{i=1}^{m} a_i f^i(\x)$ in the function space. The two goals are balanced by the parameter $B$. Hence, in summary, it learns an ensemble of auxiliary classifiers, and seeks the target classifier near the ensemble such that the risk in terms of concerned performance measure is minimized.

\subsubsection{Efficient Learning via Feature Augmentation}
Obviously, in CAPO, the auxiliary classifier $f'(\x)$ can be nonlinear classifiers such as SVM and neural network, thus the adapted classifier $f(\x)$ is nonlinear even if the delta function $f_\delta(\x)$ is linear. Empirical studies in Section~\ref{sec:exp} show that using linear delta function $f_\delta(\x)$ achieves good performance whilst keeping computational efficiency.

Consider linear delta funcion, i.e., $\Phi(\x) = \x$ and $f_\delta(\x) = \w^\top \x$, and take CAPO with multiple auxiliary classifiers for example, if we augment the original features with outputs of auxiliary classifiers, and let
\begin{eqnarray}\label{eqn-pi-mul}
\mathbf{v} = \begin{bmatrix} {\sqrt{B}} ~ \mathbf{a} \\ \w \end{bmatrix}
~~~\text{and}~~~
\x'_i = \begin{bmatrix} \frac{1}{\sqrt{B}} ~ \mathbf{f}_i \\ \x_i \end{bmatrix},
\end{eqnarray}
the adaptation problem (\ref{eqn-OP-multi}) can be written as
\begin{eqnarray}\label{eqn-OP-multi-alt}
\min_{\mathbf{v}, \xi \geq 0} && \frac{1}{2}\|\mathbf{v}\|^2 + C \xi\\
\text{s.t.} &&  \forall~\bar{y}' \in \Y^n \setminus \bar{y}: \nonumber \\
        && \mathbf{v}^\top \left[\sum_{i=1}^n y_i \x_i' - \sum_{i=1}^n y_i' \x_i'\right]
        \geq \Delta(\bar{y}, \bar{y}') - \xi \nonumber.
\end{eqnarray}
For CAPO with one auxiliary classifier, it is easy to find that there exist a constant $B$ such that the adaptation problem (\ref{eqn-OP-single}) can also be transformed into problem~(\ref{eqn-OP-multi-alt}) if we define
\begin{eqnarray}\label{eqn-pi-sgl}
\mathbf{v} = \begin{bmatrix} {\sqrt{B}} \\ \w \end{bmatrix}
~~~\text{and}~~~
\x'_i = \begin{bmatrix} \frac{1}{\sqrt{B}} ~ \mathbf{f}_i  \\ \x_i \end{bmatrix}.
\end{eqnarray}
Note that the problem~(\ref{eqn-OP-multi-alt}) is the same as that of linear \svmperf~in (\ref{eqn-svmperf}). Thus, after obtaining  auxiliary classifiers, if we augment the original data features with the outputs of auxiliary classifiers according to (\ref{eqn-pi-mul}) or (\ref{eqn-pi-sgl}), the classifier adaptation problem of CAPO~can be efficiently solved by the cutting plane algorithm in Algorithm~\ref{algo:cpa}.
Obviously, as linear \svmperf, CAPO can also handle all the performance measures based on the contingency table and AUC.

In practice, CAPO is an efficient approach for training nonlinear classifiers optimizing specific performance measures, because its both steps can be efficiently performed.
Moreover, because auxiliary classifiers can be seen as estimation of the needed classifier, it can be expected that Algorithm~\ref{algo:cpa} needs fewer iterations to converge, i.e. fewer times of solving the inference (\ref{eq:find-most}); and hence its classifier adaptation procedure can be more efficient than linear \svmperf~which searches the function space directly.
This has been validated by the experimental results in Section~\ref{sec:exp-comp}.

\section{Discussion with Related Work}\label{sec:dis}
The most famous work that optimizes performance measures is \svmperf~\cite{Joachims05}. By taking a multivariate prediction formulation, it finds the classifier in the function space directly. Our proposed CAPO works in a different manner and employs auxiliary classifiers to help find the target classifier in the function space. Furthermore, CAPO is a framework that can use different types of auxiliary classifiers. If nonlinear auxiliary classifier is used, the obtained classifier will also be nonlinear. This is very helpful, because nonlinear classifier is preferred in many applications while training nonlinear \svmperf~is computationally expensive. In summary, compared with \svmperf, CAPO can provide the needed nonlinearity whilst keeping even improving computational efficiency.

Another related work is A-SVM~\cite{Yang07}, which learns a new SVM classifier by adapting auxiliary classifiers trained in other related domains. CAPO differs from A-SVM in several aspects: 1) CAPO aims to optimize specific performance measures, while A-SVM considers hinge loss; 2) The auxiliary classifiers of CAPO are used to help find the target classifier in the function space, while A-SVM is proposed for domain adaptation~\cite{Daume-06} and it employs auxiliary classifier to extract knowledge from related domains, similar ideas can be found in \cite{Duan09}.
Generally speaking, {classifier adaptation} techniques which try to obtain a new classifier based on existed classifiers, were mainly used for domain adaptation in previous studies~\cite{Yang07,Duan09}. Here, we use classifier adaptation to optimize specific performance measures, which is quite different.

Ensemble learning is the learning paradigm which employs multiple learners to solve one task~\cite{Zhou-EMFA-12}, and it achieves state-of-the-art performance in many practice applications. In current work, the final classifier generated by CAPO is an ensemble constituting of auxiliary classifiers and the delta function. But, different from conventional ensemble methods, the component classifiers of CAPO are of two kinds and generated in two steps: first, auxiliary classifiers are trained; then a delta function which is designed to correct the decision of auxiliary classifiers is added such that the concerned performance measure is optimized.

From the feature augmentation perspective, the nonlinear auxiliary classifiers construct nonlinear features that are augmented to the original features, so that the final classifier can have nonlinear generalization performance. This is like \textit{constructive induction}~\cite{MatheusR89} which tries to change the representation of data by creating new features.

Curriculum learning~\cite{BengioLCW09} is a learning paradigm which circumvents a challenging learning task by starting with relatively easier subtasks; then with the help of learnt subtasks, the target task can be effectively solved. It was first proposed for training neural networks in~\cite{Elman93}, and is closely related to the idea of ``twice learning'' proposed in~\cite{ZhouTKDE04}, where a neural network ensemble was trained to help induce a decision tree. The study in~\cite{BengioLCW09} shows promising empirical results of curriculum learning. Our proposed CAPO is similar to curriculum learning since it also tries to solve a difficult problem by starting with relatively easier subtasks, but they are quite different because we do not provide a curriculum learning strategy.

\section{Empirical Studies}\label{sec:exp}
In this section, we perform experiments to evaluate the performance and efficiency of CAPO.

\subsection{Configuration}
The following five data sets from different application domains are used in our experiments.
\begin{itemize}
  \item \textsc{Ijcnn1}: This data set is from \textsc{Ijcnn} 2001 neural network competition (task 1), here we use winner's transformation in~\cite{Chang01ijcnn2001}.
  \item {Mitfaces}: Face detection data set from CBCL at MIT~\cite{alvira2001}.
  \item {Reuters}: Text classification data which is to discriminate the \texttt{money-fx} documents from others in the {Reuters-21578} collection.
  \item {Splice}: The task is to recognize two classes of splice junctions in a DNA sequence.
  \item \textsc{Usps*}: This data set is to classify the digits ``01234" against the digits ``56789" on the \textsc{Usps}  handwritten digits recognition data.
\end{itemize}
Table~\ref{tbl:data} summarizes the information of data sets. On each data set, we optimize 4 performance measures (accuracy, F1-score, PRBEP and AUC) so there are 20 tasks in total.
For each task, we train classifiers on training examples, and then evaluate their performances on test examples.
The experiments are run on an Intel Xeon E5520 machine with 8GB memory.

\begin{table}[!t]
\centering\scriptsize
\caption{ Data sets used in the experiments. }\label{tbl:data}
\begin{tabular}{c|cc@{\quad\quad}c}
\toprule 
{\textsc{Data set}} & \textsc{\#Feature} & \textsc{\#Train}  &  \textsc{\#Test}\\
\midrule
\textsc{Ijcnn1} & 22 & 49,990 & 91,701\\
{Mitfaces} & 361 & 6,977 & 24,045\\
{Reuters} & 8,315 & 7,770 & 3,299\\
{Splice} & 60 & 1,000 & 2,175\\
\textsc{Usps*} & 256 & 7,291 &2,007\\
\bottomrule
\end{tabular}
\end{table}


\subsection{Comparison with State-of-the-art Methods}\label{sec:exp-comp}
First, we compare the performance and efficiency of CAPO with state-of-the-art methods. Specifically, we compare three methods which can optimize different performance measures, including \svmperf, classification SVM incorporating with a cost model~\cite{MorikBJICML99}, and our proposed CAPO. Detailed implementations of these methods are described as follows.
\begin{itemize}
  \item CAPO: We use three kinds of classifiers as auxiliary classifiers, including Core Vector Machine (CVM)\footnote{\texttt{http://www.cs.ust.hk/\~{}ivor/cvm.html}. Here, we use the option ``-c 1 -e 0.001'' for all auxiliary CVMs. }~\cite{TsangJ05}, RBF Neural Network (NN)~\cite{Bishop-NN-95} and C4.5 Decision Tree (DT)~\cite{Quinlan:1993:CPM}, and corresponding CAPO's are denoted as CAPO$_{\rm cvm}$, CAPO$_{\rm nn}$ and CAPO$_{\rm dt}$, respectively.
      In CAPO$_{\rm cvm}$, the CVM is with RBF kernel $k(\x_i; \x_j) = \exp(\gamma\|\x_i - \x_j\|^2)$, where $\gamma$ is set to the default value (inverse squared averaged distance between examples), and the parameter $C$ is set to 1.
      In CAPO$_{\rm nn}$ and CAPO$_{\rm dt}$, NN and DT are implemented by WEKA~\cite{Hall:2009:WDM} with default parameters.
      Furthermore, we also implement CAPO*, which exploits all the three auxiliary classifiers.
      The parameter $C$ is selected from $C \in \{2^{-7}, \ldots, 2^7\}$ by 5-fold cross validation on training data, and the parameter $B$ of CAPO* is simply set to 1.
  \item \svmperf: We use the codes of \svmperf~ provided by Joachims~\footnote{\texttt{http://svmlight.joachims.org/svm\_perf.html}.}.
      Both linear kernel and RBF kernel are used, the corresponding methods are denoted as \svmperflin~and \svmperfrbf, respectively.
      The parameter $C$ for both methods and the kernel width $\gamma$ for \svmperfrbf~are selected from $C \in \{2^{-7}, \ldots, 2^7\}$ and $\gamma \in \{2^{-2}\gamma_0, \ldots, 2^2\gamma_0\} $ by 5-fold cross validation on training data, where $\gamma_0$ is the inverse squared averaged distance between examples.
  \item SVM with cost model: We implement the SVM with cost model with \svmlight~\footnote{\texttt{http://svmlight.joachims.org}.}, where the parameter $j$ is used to set different costs for different classes. Specifically, we use \svmlightlin and \svmlightrbf, where linear kernel and RBF kernel are used. The parameter $C$ and $j$ for both methods and the kernel width $\gamma$ for \svmlightrbf~are selected from $C \in \{2^{-7}, \ldots, 2^7\}$, $j \in \{2^{-2}, \ldots, 2^6\}$ and $\gamma \in \{2^{-2}\gamma_0, \ldots, 2^2\gamma_0\} $ by 5-fold cross validation.
\end{itemize}

\begin{table*}[!t]
\centering\scriptsize
\caption{ Performance of compared methods, where the best performance for each task is bolded and the methods that cannot be completed in 24 hours are indicated by ``N/A''. For CAPO, the raw performance of auxiliary classifier is shown in brackets following the entry of corresponding CAPO. }\label{tbl:perf} \smallskip\smallskip
\begin{tabular}{c@{~}l|c@{~}c@{~}c@{~}c|c@{~}c|c@{~}c}
\toprule 
\multicolumn{2}{c|}{\sc Task}  & CAPO$_{\rm cvm}$ & CAPO$_{\rm dt}$ & CAPO$_{\rm nn}$ & CAPO* & \svmperflin & \svmperfrbf & \svmlightlin & \svmlightrbf\\
\midrule
\multirow{4}*{\begin{sideways}{\textsc{Ijcnn1}}\end{sideways}}
& {\it Accuracy} & .9540  (.9521) & .9702 (.9702)   & .9150 (.8914)  &  \textbf{.9703}  &  .9193  &  .9658  &  N/A &  N/A \\
&  {\it F1}  &  .7620  (.7544) &  \textbf{.8473} (.8471)  &  .5753 (.2643) &  .8468  &  .5565  &  N/A &  N/A    &  N/A  \\
&  {\it PRBEP}  &  .7723 (.7376)  &  .8470 (.8364) &  .5692  (.3222) &  \textbf{.8605}  &  .6016  &  N/A &  N/A  &  N/A\\
&  {\it AUC}  &  .9607  (.8839) &  .9734 (.9464) &  .9198 (.8658)  &  \textbf{.9810}  &  .9180  &  N/A  &  N/A    &  N/A\\
\midrule
\multirow{4}*{\begin{sideways}{{Mitfaces~}}\end{sideways}}
&  {\it Accuracy}  &  \textbf{.9842} (.9839)  &  .9458 (.9302)  &  .9696 (.9067)  &  .9841&  .9727 & .9840 & .9733 &  N/A \\
&  {\it F1}  &  \textbf{.4658} (.4665) &  .1605 (.1342)  &  .2281 (.1768)  &  .4514  &  .2056  &  N/A &  .2015 &  N/A \\
&  {\it PRBEP}  &  \textbf{.5127} (.4979)  &  .1864 (.1822) &  .2500 (.1059)  &  .4873  &  .2140 & N/A &  .2309 & N/A \\
&  {\it AUC}  &  \textbf{.9148} (.9148)  &  .7991 (.7201) &  .8368 (.7979) & {.9137}  &  .8533  &  N/A &  .8450 & N/A \\
\midrule
\multirow{4}*{\begin{sideways}{{Reuters~}}\end{sideways}}
&  {\it Accuracy} &  \textbf{.9745} (.9745) & .9664 (.9660) & .9715 (.9315) & .9739 & .9727 &  .9727 & .9724 & .9721  \\
&  {\it F1}  &  .7730 (.7729) &  .6973 (.6890)  &  .7455 (.1439) & \textbf{.7731}  &  .7375 & N/A  &  .7599 &  .7540  \\
&  {\it PRBEP} & .7654 (.7709)  &  .7207 (.6871) &  .7151 (.3743) &  \textbf{.7765}  &  .7598  &  N/A &  .7709 &  .7598  \\
&  {\it AUC}  &  .9870 (.9363)  &  .9842 (.9144) &  .9868 (.8322) & .9838  &  \textbf{.9878} & N/A  &  .9872  &  .9873  \\
\midrule
\multirow{4}*{\begin{sideways}{{Splice}}\end{sideways}}
&  {\it Accuracy}  &  .8947 (.8947) & .9347 (.9347) & .9651 (.9651)  &  \textbf{.9664} & .8451  &  .8947 & .8446 & .8975\\
&  {\it F1}  &  .8955 (.8943) & .9371 (.9362) & \textbf{.9659} (.9659)  &  .9512  &  .8451  &  N/A    &  .8487  &  .8990  \\
&  {\it PRBEP}  & .8762 (.8691) & .9363 (.9355) & .9576 (.9558) & \textbf{.9584} & .8532  & N/A &  .8523  &  .9036  \\
&  {\it AUC}  &  .9457 (.8992) & .9760 (.9307) & .9836 (.9667) & \textbf{.9852}  &  .9304  & N/A &  .9267 & .9639  \\
\midrule
\multirow{4}*{\begin{sideways}{\textsc{Usps*}}\end{sideways}}
&  {\it Accuracy} & {.9691} (.9689) & .9233 (.9233) & .8520 (.7798) & .9676 & .8411 & \textbf{.9706} & N/A & N/A\\
&  {\it F1} & .9611 (.9613) & .9060 (.9053) & .8188 (.7486) & \textbf{.9617}  &  .8012  &   N/A &  N/A & N/A \\
&  {\it PRBEP} & .9500 (.9488) & .9000 (.8898) & .8195 (.7500) &  \textbf{.9573} &  .7963  & N/A & N/A & N/A \\
&  {\it AUC}  &  .9731 (.9658)  &  .9557 (.9179)  &  .9137 (.7582)  &  \textbf{.9843}  &  .9052  & N/A & N/A & N/A \\
\bottomrule
\end{tabular}
\end{table*}

\begin{table*}[!t]
\centering\scriptsize
\caption{ CPU time for parameter selection (in seconds), where the tasks not completed in 24 hours are indicated by ``N/A''. For CAPO, the CPU time for training auxiliary classifiers is not counted, and they are shown in Table~\ref{tbl:caperf-aux}. }
\smallskip\smallskip
\label{tbl:time-cv}
\begin{tabular}{cl|rrrr|rr|rr}
\toprule
\multicolumn{2}{c|}{\sc Task}  & CAPO$_{\rm cvm}$ & CAPO$_{\rm dt}$ & CAPO$_{\rm nn}$ & CAPO* & \svmperflin & \svmperfrbf & \svmlightlin & \svmlightrbf\\
\midrule
\multirow{4}*{\begin{sideways}{\textsc{Ijcnn1}}\end{sideways}}
& {\it Accuracy} & 9.3 & 11.1 & 9.9 & 11.2 & 10.0 & 96.6 & \multirow{4}*{N/A} & \multirow{4}*{N/A} \\
& {\it F1} & 9,451.5 & 9,011.5 & 14,809.3 & 6,652.8 & 12,281.3 & N/A &   &  \\
& {\it PRBEP} & 1,507.9 & 1,033.3 & 2,276.2 & 1,005.1 & 2,034.0 & N/A &  &  \\
& {\it AUC} & 88.0 & 38.0 & 124.0 & 40.6 & 112.6 & N/A &   &   \\
\midrule
\multirow{4}*{\begin{sideways}{{Mitfaces}}\end{sideways}}
& {\it Accuracy} & 9.5 & 11.2 & 23.7 & 9.0&27.2 & 27,089.3&\multirow{4}*{6,114.7}  & \multirow{4}*{N/A} \\
& {\it F1} & 465.6 & 802.5 & 1,211.5 & 379.0 & 1,189.4 & N/A & &  \\
& {\it PRBEP} & 126.9 & 183.4 & 241.6 & 119.6 & 234.4 & N/A & &  \\
& {\it AUC} & 37.7 & 48.5 & 74.0 & 30.6 & 79.3 & N/A & &  \\
\midrule
\multirow{4}*{\begin{sideways}{Reuters}\end{sideways}}
& {\it Accuracy} & 5.7 & 2.1 & 2.6 & 3.9 & 2.3 & 39,813.1 & \multirow{4}*{283.1}&\multirow{4}*{53,113.8} \\
& {\it F1} & 68.7 & 67.4 & 64.3 & 67.6 & 60.2 & N/A & &  \\
& {\it PRBEP} & 10.8 & 13.1 & 11.9 & 10.6 & 11.4 & N/A &  &  \\
& {\it AUC} & 18.9 & 8.6 & 8.7 & 3.9 & 8.1 & N/A &  & \\
\midrule
\multirow{4}*{\begin{sideways}{{Splice}}\end{sideways}}
& {\it Accuracy} & 4.0 & 484.5 & 697.1 & 2.0 & 3,602.4 & 2,187.1 & \multirow{4}*{16,297.6} & \multirow{4}*{464.2} \\
& {\it F1} & 168.2 & 592.3 & 3,373.9 & 58.4 & 10,201.5 & N/A & &  \\
& {\it PRBEP} & 11.8 & 17.0 & 27.3 & 6.8 & 82.6 & N/A & & \\
& {\it AUC} & 2.0 & 3.3 & 7.3 & 1.2 & 42.0 & N/A & & \\
\midrule
\multirow{4}*{\begin{sideways}{\textsc{Usps*}}\end{sideways}}
& {\it Accuracy} & 24.6 & 35.4 & 215.3 & 15.6 & 221.5 & 24,026.7&\multirow{4}*{N/A} & \multirow{4}*{N/A} \\
& {\it F1} & 2,199.0 & 2,605.4 & 5,429.9 & 1,514.8 & 5,225.9 & N/A &  & \\
& {\it PRBEP} & 626.2 & 566.1 & 938.9 & 404.4 & 895.2 & N/A &  & \\
& {\it AUC} & 155.6 & 139.9 & 424.3 & 76.1 & 452.5 & N/A &  & \\
\bottomrule
\end{tabular}
\end{table*}

For parameter selection, we extend the search space if the most frequently selected parameter was on a boundary.
Note that both SVM with cost model and \svmperf~are strong baselines to compare against. Lewis~\cite{Lewis01} won the TREC-2001 batch filtering evaluation by using the former, and Joachims~\cite{Joachims05} showed that \svmperf~performed better.
We apply these methods to the 20 tasks mentioned above, and report their performance. Since time efficiency is also concerned, we report the CPU time used for parameter selection. Note that if one task is not completed in 24 hours, we would stop it and mark it with ``N/A''.

\begin{table}[!t]
\centering\scriptsize
\caption{CPU time for training auxiliary classifiers (in seconds). } \smallskip\smallskip
\label{tbl:caperf-aux}
\begin{tabular}{c|rrr}
\toprule   	  	  	  	
\textsc{Data set} & CVM   & DT & NN \\
\midrule
\textsc{Ijcnn1} & 1.6 & 19.9 & 20.2\\
{Mitfaces} & 2.8 & 66.1 & 63.6 \\
{Reuters} &  2.1  & 1,689.7  & 1,771.0   \\
{Splice} & 0.1 & 0.4 & 0.9  \\
\textsc{Usps*} & 2.3 & 45.9 &  37.1 \\
\bottomrule
\end{tabular}
\end{table}

\begin{figure*}[!t]
\centering\small
  \includegraphics[width=\linewidth]{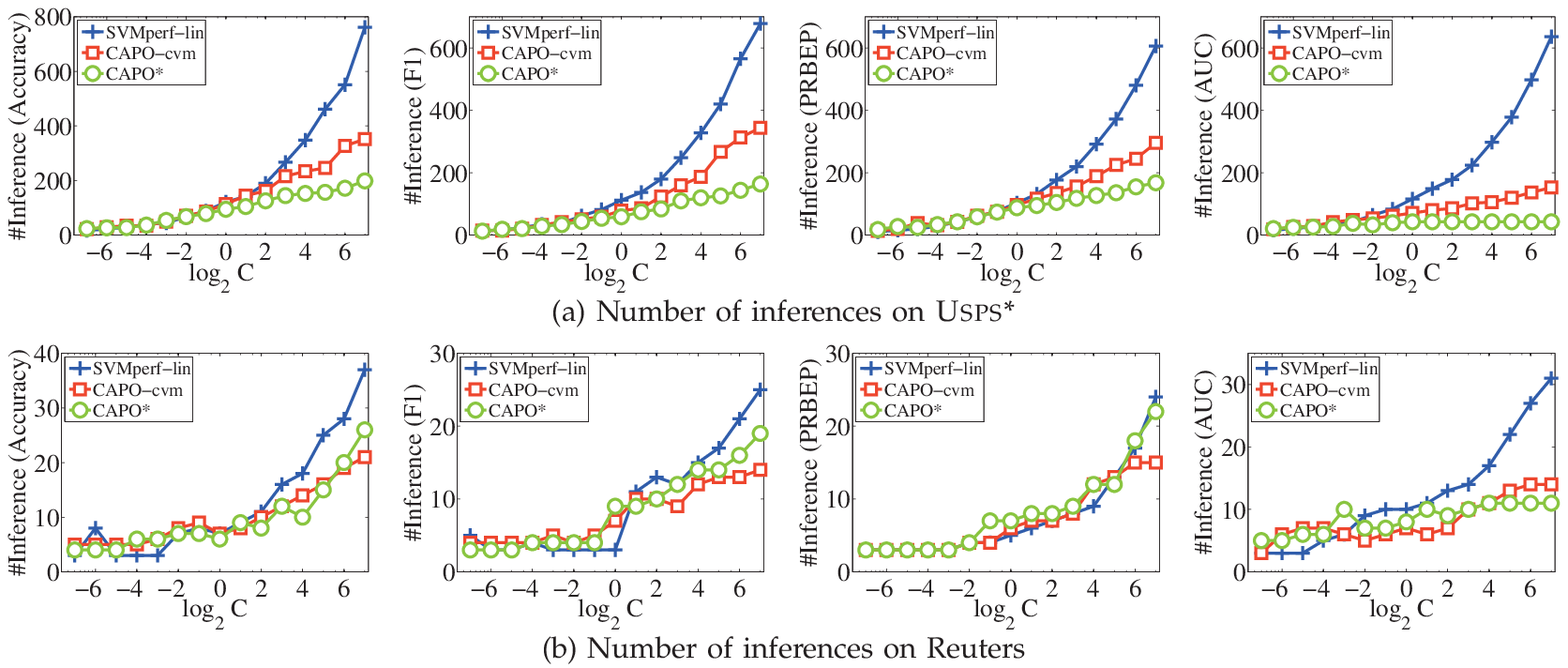}\vspace{-2mm}
\caption{ Number of inferences of the most violated constraints (\#Inference) when training \svmperflin, CAPO$_{\rm cvm}$ and CAPO* on \textsc{Usps*} and {Reuters}, where $x$-axis and $y$-axis show the $C$ values and \#Inference respectively.}
\label{fig:inference}
\end{figure*}

Table~\ref{tbl:perf} presents the performance of compared methods as well as the raw performance of auxiliary classifiers (in the brackets following the entries of corresponding CAPO methods), where the best result for each task is bolded. It is obvious that CAPO and \svmperflin~succeed to finish all tasks in 24 hours.
We can observe that CAPO achieves performance improvements over auxiliary classifiers on most tasks, and many of the performance improvements are quite large. For example, on Reuters the best AUC achieved by auxiliary classifiers is 0.9363, while CAPO methods achieve AUC higher than 0.98.  This result shows that CAPO is effective in improving the performance with respect to the concerned performance measure. More results for the case of multiple auxiliary classifiers is given in Section~\ref{sec:eodf}. 
Moreover, we could see from the results that CAPO methods perform much better than linear methods, i.e., \svmperflin~and \svmlightlin, especially when optimizing multivariate performance measures like F1-score and PRBEP.
For example, CAPO* achieves PRBEP 0.8605 but \svmperflin~achieves only 0.6016 on \textsc{Ijcnn1}; CAPO$_{\rm nn}$ achieves F1-score 0.9659, but that of \svmperflin~and \svmlightlin~are both less than  0.85 on {Splice}. This can be explained by that CAPO methods exploit the nonlinearity provided by auxiliary classifiers. Meanwhile, it is interesting that all methods achieve similar performances on { Reuters}, this coincides with the common knowledge that linear classifier is strong enough for text classification tasks. For kernelized  methods, i.e., \svmperfrbf~and \svmlightrbf, it is easy to see that they fail to finish in 24 hours on most tasks. On the smallest data set {Splice}, \svmlightrbf~succeeds to finish all tasks, its performance is better than linear methods (\svmperflin~and \svmlightlin), this can be explained that \svmlightrbf~exploits nonlinearity by using RBF kernel. Meanwhile, it is easy to see that the performances of CAPO methods especially CAPO* are superior to \svmlightrbf. This can be understood that RBF kernel may not be suitable for this data, while CAPO* exploits nonlinearity introduced by different kinds of auxiliary classifiers.
By comparing CAPO* with other CAPO methods with one auxiliary classifier, it can be found there are many cases where CAPO* performs better. This is not hard to understand because CAPO* exploits more nonlinearity by using different kinds of auxiliary classifiers.

Table~\ref{tbl:time-cv} shows the CPU time used for parameter selection via cross validation. On each data set, we employ the same auxiliary classifiers for four different measures, so the time used for training auxiliary classifiers on one data set are identical, which are shown in Table~\ref{tbl:caperf-aux}. Also, because four tasks of \svmlight~on one data set have the same cross validation process, they have the same cross validation time. From Table~\ref{tbl:time-cv} and~\ref{tbl:caperf-aux}, we can see kernelized nonlinear methods (\svmperfrbf~and \svmlightrbf) fail to finish in 24 hours on most tasks. This can be understood that the Gram matrix updating in \svmperfrbf~costs much time as described in Section~\ref{sec:svmperf}, and \svmlightrbf~has many parameters to tune.
Meanwhile, it can be found that CAPO methods are more efficient than others, even after adding the time used for training auxiliary classifiers.

Moreover, it is interesting to find that the classifier adaptation procedure of CAPO costs much less time than \svmperflin~except on { Reuters}, though it employs the later to solve the adaptation problem.
For example, when optimizing F1-score on {Splice}, CAPO* consumes only 58.4 seconds for cross validation while \svmperflin~costs more than 10,000 seconds.
To understand this phenomenon, we record the number of inferences of the most violated constraint (i.e. solving the argmax in (\ref{eq:find-most}) when training \svmperflin, CAPO$_{\rm cvm}$ and CAPO*. Concretely, on two representative data sets {Reuters} and  {\sc Usps*}, the number of inferences under different $C$ values are recorded and Figure~\ref{fig:inference} shows the results. From Figure~\ref{fig:inference} (a), we can find that on {\sc Usps*}, CAPO* and CAPO$_{\rm cvm}$ have fewer inferences than \svmperflin, especially when $C$ is large. Since the training cost of Algorithm~\ref{algo:cpa} is dominated by the inference, the high efficiency of CAPO* and CAPO$_{\rm cvm}$ is due to fewer number of inferences. This can be understood by that auxiliary classifiers provide estimates of the target classifier and CAPO searches them, while \svmperflin~searches in the whole function space.
On {Reuters} where three methods have similar time efficiency, we can find from Figure~\ref{fig:inference} (b) that the numbers of inferences are small and similar. This can be understood that linear classifier is strong enough for text classification tasks.
Moreover, the adaptation procedure of CAPO* is more efficient than CAPO$_{\rm cvm}$, and Figure~\ref{fig:inference} (a) also shows CAPO* has fewer number of inferences. This indicates that it may be easier to find the target classifier by using multiple auxiliary classifiers, coinciding with the fact that an ensemble can provide better estimate of the target classifier.

Therefore, we can see that the auxiliary classifiers not only inject nonlinearity, but also make the classifier adaptation procedure more efficient.


\subsection{Effect of Delta Function}\label{sec:eodf}
To show the effect of adding delta function on auxiliary classifiers, we compare the performance of CAPO with that of the weighted ensemble of auxiliary classifiers which does not include a delta function.
In detail, we train five CVMs as auxiliary classifiers due to its high efficiency. Each CVM is with one of the following five kernels: 1)  RBF kernel $k(\x_i; \x_j) = \exp(\gamma\|\x_i - \x_j\|^2)$; 2) polynomial kernel $k(\x_i; \x_j) = (\gamma\x_i^\top \x_j + c_0)^d$; 3) Laplacian kernel $k(\x_i; \x_j) = \exp(\gamma\|\x_i - \x_j\|)$; 4) inverse distance kernel $k(\x_i; \x_j) = \frac{1}{\sqrt{\gamma}\|\x_i - \x_j\| + 1}$; and 5) inverse squared distance kernel $k(\x_i; \x_j) = \frac{1}{{\gamma}\|\x_i - \x_j\|^2 + 1}$, where all kernels are with default parameters ($c_0 = 0$ and $d = 3$ in the polynomial kernel, $\gamma$ is the inverse squared averaged distance between examples in all kernels).
Then, CAPO employs these five CVMs as auxiliary classifiers, and the weighted ensemble learns a set of weights to combine them such that the empirical risk is minimized. Both methods select $C$ from $\{2^{-7}, \ldots, 2^7\}$ by 5-fold cross-validation on training data, and $B$ of CAPO is fixed to 1.

\begin{table}[!t]
\centering\scriptsize
\caption{ Performance comparison between CAPO and weighted ensemble, where both methods exploit five CVMs with different kernels. }\smallskip\smallskip
\label{tbl:caperf-diff}
\begin{tabular}{c@{~}l|cc}
\toprule
\multicolumn{2}{c|}{\sc Task}  &
CAPO  & Ensemble \\
\midrule
\multirow{4}*{\textsc{Ijcnn1}}
& {\it Accuracy} & .9712 & .9632 \\
& {\it F1} & .8438 & .8439 \\
& {\it PRBEP} & .8472 & .8000 \\
& {\it AUC} & .9892 & .9837 \\
\midrule
\multirow{4}*{{Mitfaces}}
& {\it Accuracy} & .9842 & .9837\\
& {\it F1} & .4563 & .4446 \\
& {\it PRBEP} & .4831 & .4767 \\
& {\it AUC} & .9097 & .9097 \\
\midrule
\multirow{4}*{{Reuters}}
&  {\it Accuracy}  &  .9715  &  .9715   \\
&  {\it F1}  &  .7429  &  .7181   \\
&  {\it PRBEP}  &  .7598  &  .7318   \\
&  {\it AUC}  &  .9847  &  .7979   \\
\midrule
\multirow{4}*{{Splice}}
& {\it Accuracy} & .8952 & .8938 \\
& {\it F1} & .9024 & .9024 \\
& {\it PRBEP} & .9010 & .8912 \\
& {\it AUC} & .9486 & .9022 \\
\midrule
\multirow{4}*{\textsc{Usps*}}
& {\it Accuracy} & .9706 & .9701 \\
& {\it F1} & .9659 & .9644 \\
& {\it PRBEP} & .9634 & .9622 \\
& {\it AUC} & .9823 & .9705 \\
\bottomrule
\end{tabular}
\end{table}

Table~\ref{tbl:caperf-diff} presents the performances of two methods. It can be seen that CAPO achieves better performance than the weighted ensemble. For example, the weighted ensemble achieves PRBEP 0.8000 on \textsc{Ijcnn1} while CAPO achieves 0.8472; the weighted ensemble achieves AUC 0.9022 but CAPO achieves 0.9486 on {Splice}. Noting that their difference is that CAPO exploits the delta function, we can see that by adding the delta function, CAPO achieves performance improvement w.r.t. concerned performance measure.

\begin{figure*}[!t]
\centering 
\includegraphics[height=0.15\linewidth]{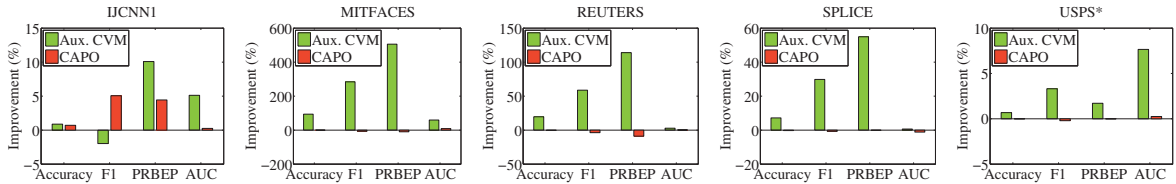}
\caption{ Comparison between relative improvement of averaged performance of auxiliary classifiers and relative performance improvement of CAPO after auxiliary classifier selection.}
\label{fig:imp}
\end{figure*}

\subsection{Effect of Auxiliary Classifier Selection}\label{sec:exp-select}
In above experiments, we directly use common learning algorithms to train auxiliary classifiers, it is obvious that these auxiliary classifiers are not specially improved according to the concerned performance measure. Then, a straightforward question is how CAPO performs if the auxiliary classifiers are specially improved w.r.t. the concerned performance measure, or in other words how CAPO performs if we train auxiliary classifiers according to the concerned performance measure. Subsequently, we perform experiments to answer this question.
Specifically, rather than training five CVMs with five different kernels with default parameters, we train a set of fifty CVMs and select five from them as auxiliary classifiers based on the concerned performance measure.
In detail, these fifty CVMs are trained by independently using the five kernels mentioned above, and the parameter $\gamma$ for each kernel is set as $ \gamma = 1.5^\theta\gamma_0$, where $\theta \in \{-0.5, 0, 0.5, \ldots, 4\}$ and $\gamma_0$ is the default value, and then five CVMs which performs best in terms of the concerned performance measure are selected as auxiliary classifiers. For example, if we want to train classifier optimizing F1-score, then the five CVMs which achieves the highest F1-score are selected. As above, we choose the parameter $C \in \{2^{-7}, \ldots, 2^7\}$ by 5-fold cross validation and fix $B$ to be 1.

On each task, we compute the relative improvement of the averaged performance of auxiliary classifiers and that of obtained CAPO, and report them in Figure~\ref{fig:imp}. The relative performance improvement is computed as the performance improvement caused by the auxiliary classifier selection divided by the performance before selection. From Figure~\ref{fig:imp}, it is easy to see that although the averaged performance of auxiliary classifiers improves a lot after selection, yet the performance of CAPO keeps similar in most cases, and even degrades in some cases. This may suggest that it is enough to use common CVMs as auxiliary classifiers, and it is not needed to specially design auxiliary classifiers according to the target performance measure. This can be explained that the auxiliary classifiers are used to provide approximate solutions to the problem, which are combined and further refined by the delta function to obtain the final solution, thus actually these approximate solutions are not required to be very accurate.
Moreover, it is obvious that with respect to time efficiency, CAPO with auxiliary classifier selection has no superiority over the original one, especially after counting the time used for training fifty auxiliary CVMs.

\begin{figure*}[!ht]
  \centering\small
  \includegraphics[width=\linewidth]{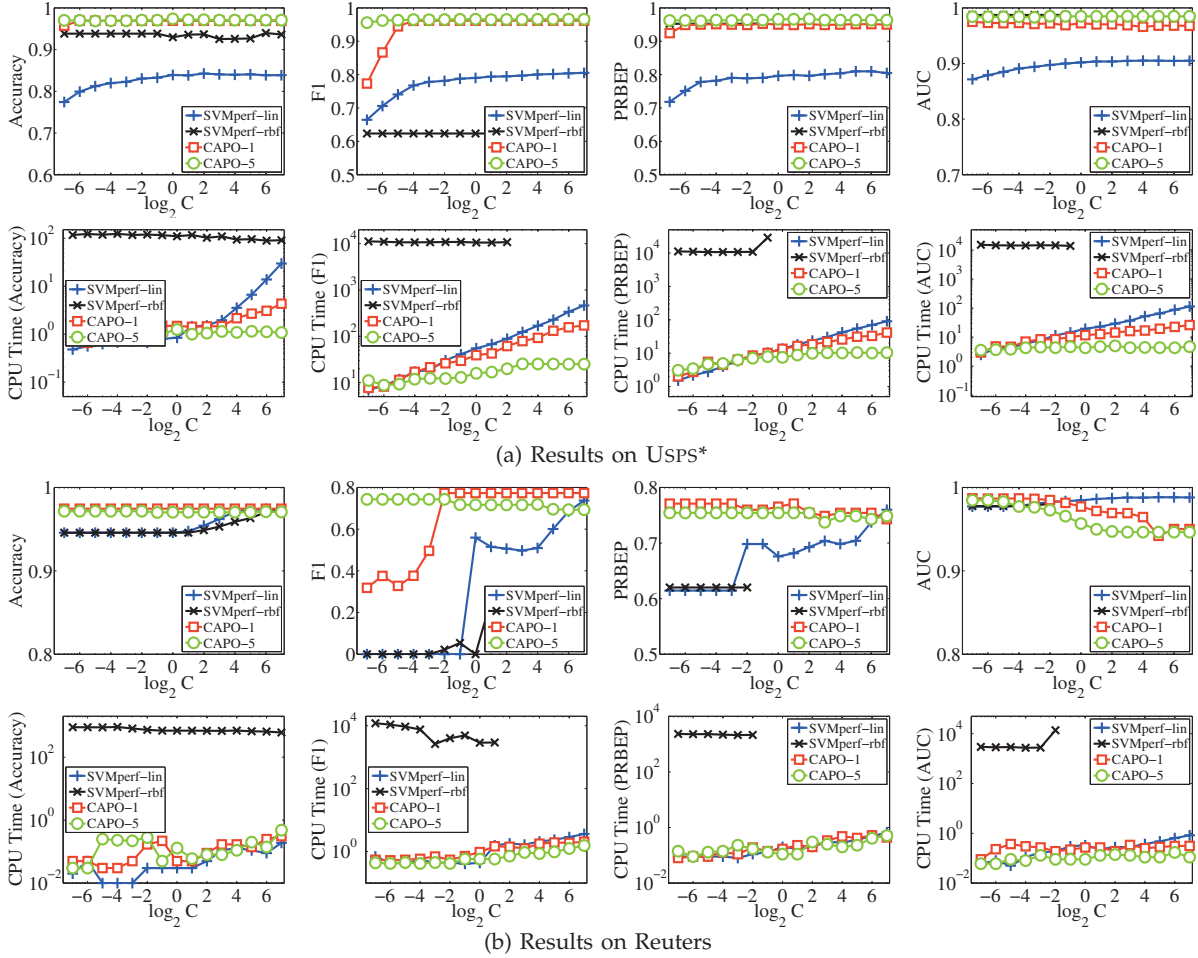}\\ \vspace{-2mm}
  \caption{ Performance and CPU time (in seconds) with different $C$'s, (a) on \textsc{Usps*}; (b) on {Reuters}. Each subfigure shows performance in the 1st row and corresponding CPU time in the 2nd row. }\label{fig:c}
\end{figure*}

\begin{figure*}[!ht]
  \centering \small
  \includegraphics[width=\linewidth]{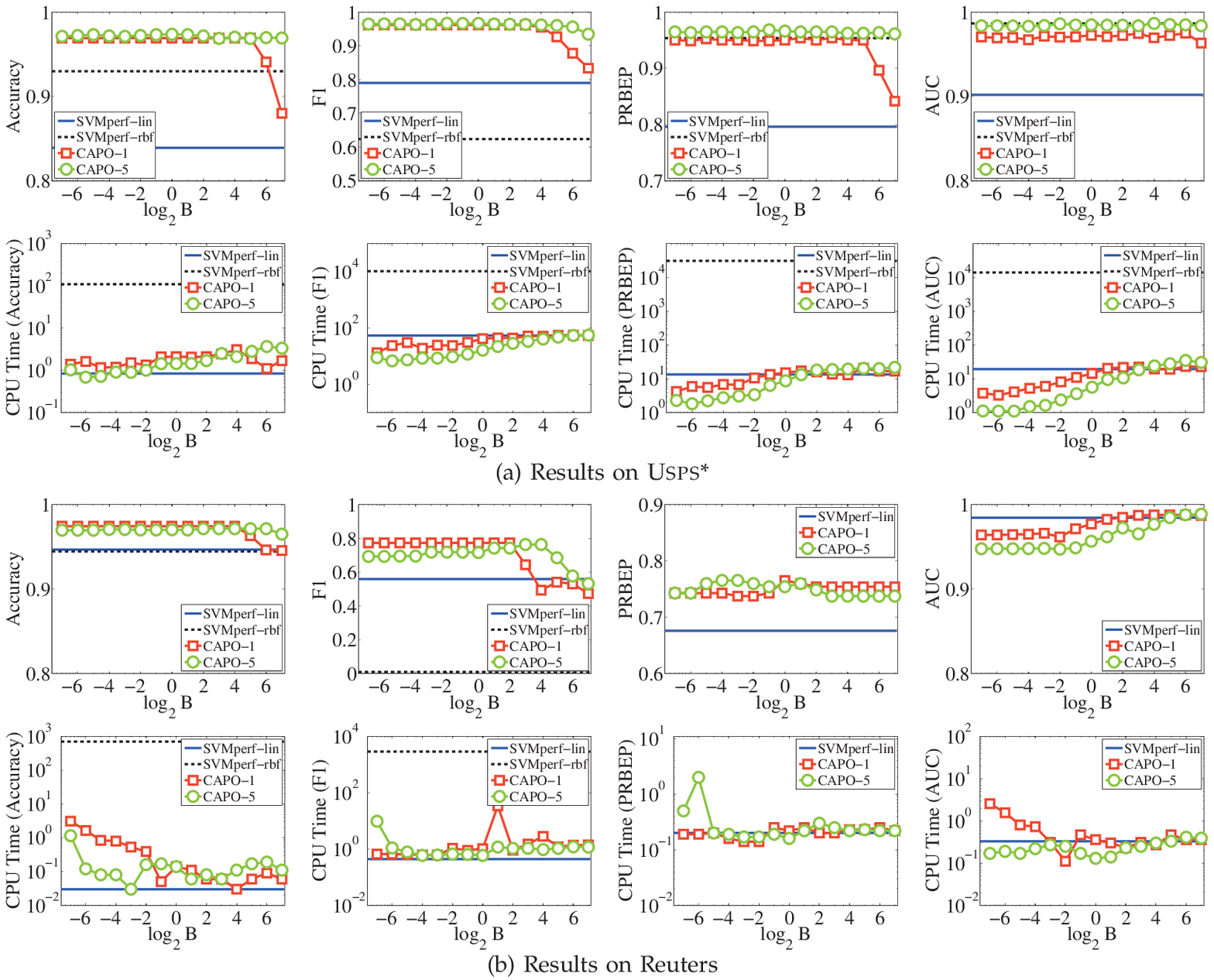}\vspace{-2mm}
  \caption{ Performance and CPU time (in seconds) with different $B$'s: (a) on \textsc{Usps*}, (b) on {Reuters}. Each subfigure shows performance in the 1st row and corresponding CPU time in the 2nd row. }\label{fig:b}
\end{figure*}

\subsection{Parameter Sensibility}
To study the impact of parameters, we perform experiments on two medium-sized data sets \textsc{Usps*} and {Reuters}. The two data sets are representative, since nonlinear classifiers perform well on \textsc{Usps*} while linear classifiers work well on {Reuters}. We study the performance and time efficiency of CAPO$_1$ and CAPO$_5$ under different $C$ and $B$ values, where CAPO$_1$ uses one auxiliary CVM with RBF kernel and CAPO$_5$ uses five auxiliary CVMs with five different kernels as above, all kernels are with default parameters.

First, we vary $C$ within $\{2^{-7}, 2^{-6},\ldots,2^7\}$ and fix $B$ to be $1$. For comparison, we also train \svmperflin~and \svmperfrbf~with the same $C$'s.  Figure~\ref{fig:c} shows the results. %
It can be found that \caperfone~and \caperffive~generally outperform \svmperf~at different $C$'s, except that \svmperfrbf~achieves comparable performance as \caperf~for PRBEP and AUC on \textsc{Usps*} and \svmperflin~performs better for AUC ~at large $C$'s on {Reuters}. With respect to time efficiency, \caperfone, \caperffive~and \svmperflin~cost comparable CPU time, which is much less than \svmperfrbf. Moreover, \caperfone~and \caperffive~scales better when $C$ increases, and they are more efficient than \svmperflin~at large $C$'s. Moreover, it is easy to find that our methods, especially CAPO$_5$, are more robust with $C$.

Second, we vary $B$ within $\{2^{-7}, 2^{-6},\ldots,2^7\}$ with fixed $C = 1$ for CAPO$_1$ and CAPO$_5$. As comparisons, \svmperflin~and \svmperfrbf~are trained  $C = 1$. The results are shown in Figure~\ref{fig:b}, where \svmperflin~and \svmperfrbf~are illustrated as straight lines because they do not have the parameter $B$. In general, \caperfone~and \caperffive~achieve better performance at different $B$'s in most cases, except for AUC on {Reuters}. Also, \caperfone~and \caperffive~have comparable efficiency with \svmperflin, which is much better than \svmperfrbf. We can see that our methods are  quite robust to parameters $B$, and comparatively speaking, \caperffive~is  more robust than \caperfone.

Thus, we can see that our methods, especially CAPO$_5$, are robust to $B$ and $C$. Comparatively speaking, \caperffive~is more robust and more efficient than \caperfone, this verifies our previous results.


\begin{figure*}[!t]
  \centering\small
  \includegraphics[width=\linewidth]{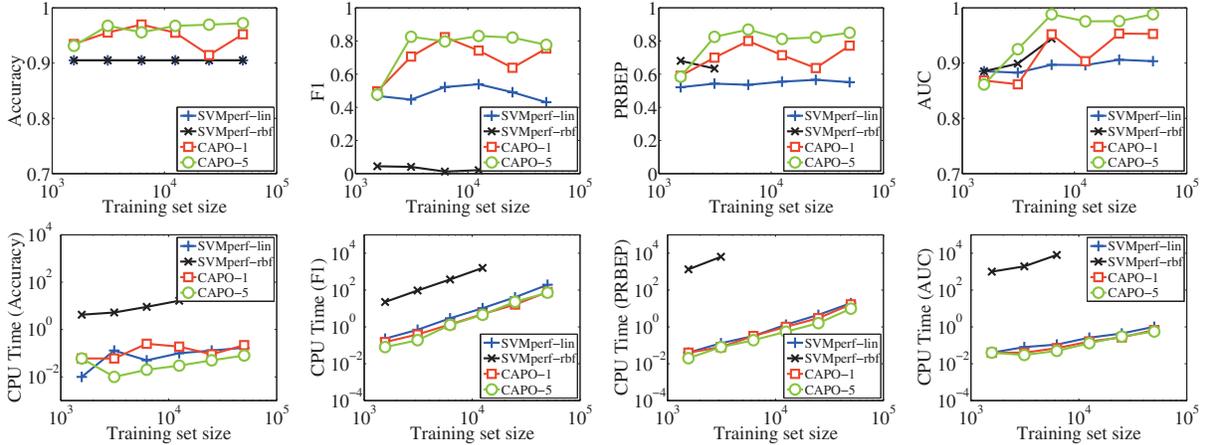}\vspace{-2mm}
\caption{Performance (1st row) and CPU time (2nd row; in seconds) with different training set sizes on \textsc{Ijcnn1}. }\label{fig:s}
\end{figure*}

\subsection{Scalability w.r.t. Training Set Size}
To evaluate scalability of \caperf, we perform experiments on the largest data set \textsc{Ijcnn1}. We first train \caperfone~and \caperffive~using $\{1/32, 1/16, 1/8, 1/4, 1/2, 1\}$ of all training examples, and then evaluate them on test examples. As comparisons, \svmperflin~and \svmperfrbf~are also trained under the same configuration. In this experiment, we simply fix both the parameters $B$ and $C$ to be 1. We report performance of compared methods and the corresponding used CPU time.

Figure~\ref{fig:s} shows the results of the achieved performance and the corresponding running time in first and second row respectively. As we can see, all methods scale well except that \svmperfrbf~has to be terminated early when the training set size increases. Moreover, compared with \svmperflin, it is easy to see that \caperffive~achieves better performance but costs less time at every training set size.

\subsection{Summary}
Based on above empirical studies, we can see that CAPO is an effective and efficient approach to training classifier that optimizes performance measures. Compared with \svmperf~and SVM with cost model, it can achieve better performances at lower time costs. As well, it has been shown that CAPO is robust to parameters and scales well w.r.t. the training data size. For practical implementation, training auxiliary classifiers by optimizing accuracy is a good choice, because many efficient algorithms have been developed in the literature, and the experiments in Section~\ref{sec:exp-select} suggest that using auxiliary classifiers with higher target performances does not show significant superiority, especially when tuning auxiliary classifiers costs much time. Meanwhile, it can be better to use multiple diverse auxiliary classifiers.

\section{Conclusion and Future Work}\label{sec:conclusion}
This paper presents a new approach CAPO to training classifier that optimizes specific performance measure. Rather than designing sophisticated algorithms, we solve the problem in two steps: first, we train auxiliary classifiers by taking existing off-the-shelf learning algorithms; then these auxiliary classifiers are adapted to optimize the concerned performance measure. We show that the classifier adaptation problem can be formulated as an optimization problem similar to linear \svmperf~and can be efficiently solved.
In practice, the auxiliary classifier (or ensemble of auxiliary classifiers) benefits CAPO in two aspects:
\begin{enumerate}
  \item By using nonlinear auxiliary classifiers, it injects nonlinearity that is quite needed in practical applications;
  \item It provides an estimate of the target classifier, making the classifier adaption procedure more efficient.
\end{enumerate}
Extensive empirical studies show that the classifier adaptation procedure helps to find the target classifier for the concerned performance measure. Moreover, the learning process becomes more efficient than linear \svmperf, due to fewer inferences in CAPO.

In this work, linear delta function is used for classifier adaptation. Although it achieves good performances, an interesting and promising future work is to exploit nonlinear delta function for this problem.

\bibliography{capo-s}
\bibliographystyle{plain}

\end{document}